\documentclass[runningheads]{llncs}

\usepackage[hidelinks]{hyperref}
\usepackage[T1]{fontenc}
\usepackage{graphicx}
\usepackage{float}
\usepackage[
backend=biber,
style=numeric,
maxnames=50,
]{biblatex}

\usepackage[rgb]{xcolor}
\usepackage{pgfplots}
\pgfplotsset{compat=1.12}

\begin{document}
\title{Supervised Fine-Tuning LLMs to Behave as Pedagogical Agents in Programming Education}
%
%
\author{Emily Ross\inst{1, *} \and
Yuval Kansal\inst{2, *} \and
Jake Renzella\inst{1} \and
Alexandra Vassar\inst{1} \and
Andrew Taylor\inst{1}}

\authorrunning{Ross \& Kansal et al.}

\institute{University of New South Wales, Sydney, Australia 
\email{\{emily.ross1, jake.renzella, a.vassar, andrewt\}@unsw.edu.au} \and
Princeton University, Princeton, NJ 08544, USA 
\email{yuvalkansal@princeton.edu} \\
\medskip
\textsuperscript{*}Equal first authors}

\maketitle              

\begin{abstract}
    Large language models (LLMs) are increasingly being explored in higher education, yet their effectiveness as teaching agents remains underexamined. In this paper, we present the development of GuideLM, a fine-tuned LLM designed for programming education. GuideLM has been integrated into the Debugging C Compiler (DCC), an educational C compiler that leverages LLMs to generate pedagogically sound error explanations. Previously, DCC relied on off-the-shelf OpenAI models, which, while accurate, often over-assisted students by directly providing solutions despite contrary prompting.

    To address this, we employed supervised fine-tuning (SFT) on a dataset of 528 student-question/teacher-answer pairs, creating two models: GuideLM and GuideLM-mini, fine-tuned on ChatGPT-4o and 4o-mini, respectively. We conducted an expert analysis of 400 responses per model, comparing their pedagogical effectiveness against base OpenAI models. Our evaluation, grounded in constructivism and cognitive load theory, assessed factors such as conceptual scaffolding, clarity, and Socratic guidance.

    Results indicate that GuideLM and GuideLM-mini improve pedagogical performance, with an 8\% increase in Socratic guidance and a 58\% improvement in economy of words compared to GPT-4o. However, this refinement comes at the cost of a slight reduction in general accuracy. While further work is needed, our findings suggest that fine-tuning LLMs with targeted datasets is a promising approach for developing models better suited to educational contexts.


\end{abstract}

\section{Introduction}
Novice programmers face various challenges when learning a new programming language, from struggling to understand and interpret compiler error messages, to fundamental misunderstanding of concepts and skills \cite{Becker2016EffectiveStudents, Kohn2019ThePython, Karvelas2020TheBehavior, Becker2019CompilerResearch}. To alleviate these challenges, educational tools, such as the Debugging C Compiler (DCC) aim to produce less cryptic and more helpful error messages for novices \cite{Taylor2023FoundationsCompiler}. Integrations to the tool leverage large language models (LLMs) with compiler context to generate bespoke, novice-friendly error explanations \cite{Taylor2024DccModels}. 
LLMs' ability to produce code from simple prompts \cite{Chen2021EvaluatingCode} has also driven a surge in student adoption of tools like ChatGPT, Claude, and Llama for assignments and everyday coding tasks \cite{Denny2024ComputingAI}. However, there are serious concerns that the indiscriminate use of such tools may violate the traditional principles of pedagogy that state that students learn by doing \cite{Ben-Ari2001} over time and with effort \cite{Sweller2023CognitiveLearn}. The authors of DCC \cite{Taylor2024DccModels} report that LLMs ignore prompts to not provide complete solutions to queries in 48\% of cases, further indicating that prompt engineering techniques may not be able to resolve some of these important pedagogical concerns. Violating pedagogical principles may lead to student overreliance on these tools, which may hinder their ability to develop critical thinking skills effectively \cite{Razafinirina2024PedagogicalChallenges, Sullivan2023ChatGPTLearning}, develop skills essential to analyse and interpret generated code, and assess potential security vulnerabilities present in LLM-generated code \cite{Liu2024FromPrompting, Black2024BalancingModels}.

To address these challenges, we performed supervised fine-tuning to develop GuideLM --- a pedagogically sound LLM designed to provide human tutor-like assistance with C/C++ syntax, coding style, and common challenges faced by novice programmers. GuideLM aims to provide Socratic guidance while preserving an \textit{economy of words} to guide students towards understanding, without overhelping. The training dataset included 528 student-question/expert tutor-answer pairs filtered from a course forum dataset of 13,000 CS1 responses.
In this paper, we attempt to answer the following research questions:
 
 \begin{itemize}
    \item[\textbf{RQ1}] How does GuideLM perform compared to foundational models in providing tutor-like responses to student programming queries?
    \item[\textbf{RQ2}] How does the pedagogical fine-tuning process affect the overall performance and accuracy of models?
 \end{itemize}

Additionally, we present the following key contributions in this paper: 
\begin{enumerate}
    \item A methodology for developing the dataset through manual and automatic filtering to ensure high quality inputs for fine-tuning foundational models.
    \item A pedagogically aligned model capable of providing tutor-like help. 
    \item A manual evaluation of GuideLM's LLM-generated responses across multiple questions conducted by subject experts.
\end{enumerate}

\section{Background}
As LLM tools gain prominence across various sectors, exploring their potential in education has become even more crucial. 

\subsection{Large Language Model-Based Tools in Computing Education}
In their review of LLM-based systems for education, Garcia-Mendez et al.~\cite{Garcia-Mendez2024ATutors} identified the most common applications of these tools, including assistance with question generation, grading student work, and code correction and explanation. In CS1 contexts, applications include solving simple coding problems \cite{Denny2023ConversingLanguage, Finnie-Ansley2022TheProgramming, Denny2024DesirableEducation, Finnie-Ansley2023MyExercises, Wermelinger2023UsingProblems}, generating explanations for code \cite{MacNeil2022ExperiencesE-Book, Liu2024TeachingEducation}, and for assistance with resolving compilation errors \cite{Taylor2023DccModels, LeeSolano2024DCCInterface, Liffiton2023CodeHelp:Classes}. Google have also recently introduced the \textit{pedagogical instruction following} (LearnLM) paradigm which allows educators to add contextual system prompts to achieve desired behaviour of their Gemini-powered LLM \cite{LearnLMTeam2024LearnLM:Learning}. The paradigm incorporates prompt engineering, RLHF, and Fine-Tuning to improve pedagogical behaviour, with raters preferring it 31\% over GPT-4o and 11\% over Anthropic's Claude 3.5 Sonnet \cite{LearnLMTeam2024LearnLM:Learning}.

At Harvard University, the CS50 bot was developed and deployed for the university's introductory programming course, utilising GPT-4 to assist students by answering queries on the forum, and to help debug code \cite{Liu2024TeachingEducation}. The bot was also available as part of an Integrated Development Environment (IDE), providing support beyond normal business hours. Marketed as a complement to human instruction rather than a replacement, the bot was tested by thousands of students over the course of a year. Although it utilised a Retrieval-Augmented Generation (RAG) pipeline and integrated course materials such as lecture notes and recordings, it lacked query-specific insights into code errors. The effectiveness of RAG depends on the availability of high-quality content, which may not always be present for general queries or across all courses.

To address scalability concerns in large, diverse introductory courses, a virtual teaching assistant (TA) was built using the LangChain \cite{Mavroudis2024LangChain} framework for an introductory computing course \cite{Liu2024BeyondScience}. Powered by OpenAI’s GPT-3.5, the system demonstrated accuracy comparable to human teaching assistants, but received higher ratings for clarity and engagement; however, often overwhelmed novices with its information density\cite{Liu2024BeyondScience}. 


A significant challenge of LLM tool use in education is the harmful effect it may have on student learning outcomes \cite{Denny2024ComputingAI}. The propensity with which LLM-powered tools provide direct answers despite prompting to the contrary is a key example \cite{Taylor2024DccModels}. In introductory programming, developing computational skills is instrumental in fostering higher-order thinking and effective problem solving \cite{Loksa2022MetacognitionUse, Prather2024TheProgrammers}. Student reliance on LLM-powered code generation may hinder their ability to develop such crucial skills if they are simply given the answer \cite{Denny2024ComputingAI}. 



\subsection{Large Language Model Fine-Tuning}
While LLMs perform well on natural language understanding and generation tasks \cite{Touvron2023LlamaModels}, more work is needed in the area of personalisation and alignment \cite{Touvron2023LlamaModels, Bubeck2023SparksGPT-4}. Fine-tuning has proven highly effective for personalisation tasks, and involves adjusting the model weights of a pre-trained model to better fit a new dataset. 

Fine-tuning techniques such as Supervised Fine-Tuning (SFT) via low-rank adaptation (LoRA) \cite{Hu2021LoRA:Models} and parameter-efficient fine-tuning (PEFT) \cite{Xu2023Parameter-EfficientAssessment} are commonly used to incorporate new knowledge into base models, offering significant potential for adapting to specific response styles \cite{Zhu2024LifelongRecommendation}. These approaches present the current state-of-art to tailor models to deliver tutor-like responses given an appropriate fine-tuning dataset. One of the more commonly used fine-tuning methods, SFT, adapts a pre-trained model to a specific task by taking a labelled dataset as input constructed for the intended task \cite{Parthasarathy2024TheOpportunities}. To be effective, a significant amount of raw data and resources are required to construct and label SFT datasets. However, the payoff is better accuracy and coherence in domain-specific applications \cite{Parthasarathy2024TheOpportunities}. The success of Codex \cite{Chen2021EvaluatingCode}, a GPT language model which has been fine-tuned on publicly available GitHub code, has shown that pre-trained models can be successfully adapted to source code. Simple fine-tuning on downstream tasks, such as closed-book question answering has shown dramatic improvement in code generation \cite{Lomshakov2023Fine-TuningSnippets}. Using code review automation tasks, another study showed the fine-tuned approach provided a 73\% improvement in providing code revisions, when comparing the effectiveness of zero-shot learning on a fine-tuned model, using GPT-3.5, against the base model \cite{Pornprasit2024Fine-tuningAutomation}.

These findings show promise in fine-tuning for educational purposes across code debugging tasks. Another study using a coding dataset of over 200 questions, covering both coding and text-based topics, examined updating knowledge through fine-tuning and reported an increase in the effectiveness of GPT-3.5-turbo, GPT-4o, and GPT-4o mini \cite{Wu2024FineTuneBench:LLMs}. The findings showed that the fine-tuned GPT-4o mini and GPT-3.5-turbo achieved nearly 100\% accuracy in reproducing answers to coding questions, while the fine-tuned GPT-4o reached 94\% accuracy \cite{Wu2024FineTuneBench:LLMs}. On text-based questions, all fine-tuned models demonstrated at least a 150\% improvement when responding to rephrased inputs \cite{Wu2024FineTuneBench:LLMs}. These results highlight the effectiveness of OpenAI's fine-tuning approach in both code and text-based applications, and aligns with this project's goals for educational enhancement.

\section{Modelling}
With the goal of improving pedagogically-sound programming error explanations provided to students in the DCC suite of programming tools, this section describes the data source, foundation model selection, and fine-tuning techniques which comprise  the development of GuideLM. The design goals were inspired by educational theories such as constructivism and cognitive load theory, as well as metrics synthesised from previous work in this space \cite{Taylor2024DccModels}. The design goal properties are described in \autoref{tab:my_label}.

\begin{table}[H]
\caption{Modelling design goal properties. \footnotesize \textit{*metric adopted from existing work.}}
    \begin{tabular}{|l|l|p{78mm}|}
        \hline Key & Property & Description\\\hline
        C1 & Conceptually Accurate* & Is the generated response conceptually correct?\\\hline
        C2 & Inaccuracy Present* & Is there inaccurate information present in response?\\\hline
        C3 & Suggestions Correct* & Is the provided guidance technically correct, resulting in being able to solve the problem?\\\hline
        C4 & Relevant to the Error* & Is the generated response relevant to the error?\\\hline
        C5 & Relevant to the Novice* & Is the generated response relevant to the novice?\\\hline
        C6 & Complete Explanation* & Is the provided explanation complete, including all critical information?\\\hline
        C7 & Overhelpful & Is the response provided overhelpful?\\\hline
        C8 & Economy of Words & Is the error described with as few words as possible to convey all necessary information?\\\hline
        C9 & Socratic Guidance & Does the response give students concepts to think about, rather than providing solutions explicitly?\\\hline
    \end{tabular}
    
    \label{tab:my_label}
\end{table}

\begin{figure}[ht]
    \vspace{-0.7cm}
    \includegraphics[width=\textwidth]{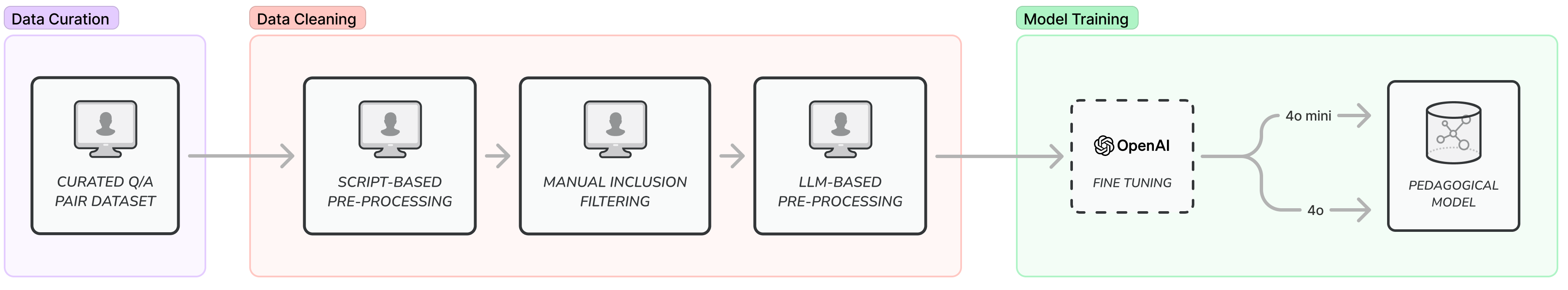}
    \vspace{-0.7cm}
    \caption{GuideLM model fine-tuning process, depicting training data curation, data pre-processing, and fine-tuning output.}\label{fig_1}
\end{figure}

\subsection{Data Source}
To accurately emulate the tutor response style, we extracted questions and answers from the university's internal course forum. On this platform, trained teaching assistants, compensated for their work, respond to students' course and code-related enquiries. Fifteen unique courses have been hosted on this forum platform. This data was accessed after gaining university ethics approval, and gaining opt-out consent from individual lecturers. Ten courses opted-out and were not included in the forum dataset. The remaining courses contributed an estimated 129,000 question-answer pairs, scraped from the forum via HTTP requests. Around 13,000 of these were from CS1 course iterations, therefore shifting the focus to improving educational outcomes for CS1 application. Exact distributions of gender, race and other demographic variables is unknown; however, the dataset broadly represents a diverse cohort of Australian higher education students.

\subsection{Model and Environment}
OpenAI was selected as the initial fine-tuning environment. At the time of work, OpenAI's offerings presented the most simple and user-friendly capabilities for fine-tuning experiments. The selected foundation models were the current state-of-the-art models, GPT-4o and GPT-4o mini, with reasoning and logic capabilities similar to those of other non-open source models \cite{Zhao2024DocMath-Eval:Documents}. A training dataset in JSONL format was required. The platform provided a straightforward, web-based method that did not require the acquisition of hardware resources or GPUs. The total cost to develop both GuideLM models was US\$250.

\subsection{Data Processing}
Prior to conducting the supervised fine-tuning with the collected course dataset, a number of pre-processing steps were required which are outlined below:

\subsubsection{1. Script-Based Pre-processing}
A Python script was created to streamline the CS1 question-answer dataset. The CS1 data was loaded into a Pandas dataframe and then each entry was filtered for any personally identifiable information or other irrelevant information that may hinder the educational outcomes of the dataset. This included newlines, email addresses, student ID numbers, URLs, and student or tutor name references. Course templates, used in many course forums to streamline question quality, were also identified and removed to reduce redundant information. Pairs containing a question or answer less than 9 or 2 characters respectively were also discarded to encourage context richness and pedagogical soundness in the dataset. A variety of techniques were used to detect the dirty data. These included using regular expressions to identify ID numbers and email addresses, and blacklists of known course templates, which were removed from the question or answer string if matched.

\subsubsection{2. Manual Filtering}
Following pre-processing scripts and manual filtering was designed to ensure dataset quality. We identified that many entries did not include appropriate problem context, or concerned non-programming contexts such as course administration questions, which could be detrimental to the technical value of the dataset. Five senior CS1 teaching assistants were tasked with assessing a sample of 500 question-answer pairs each, picked at random from the initial set of 13,000 entries, according to the following criteria:
\begin{itemize}
    \item \textbf{Good quality}: a correct and helpful response.
    \item \textbf{Self-contained}: a complete and definitive answer to the question that should not refer to other sources.
    \item \textbf{Not over-helpful}: provides polite suggestions or encouragement, but not a complete code fix.
    \item \textbf{Formal} in tone and not dismissive.
    \item \textbf{Demonstrative code blocks only}: no corrected, fully-working solutions to the student's problem or given code.			
    \item \textbf{Unidentifiable} information.			
    \item \textbf{Does not include assessment details}.				
    \item Focus on understanding the \textbf{C language}, common bugs and style.
\end{itemize}

From this criteria, each pair was then assigned a category:
\begin{itemize}
    \item \textbf{Yes}: All criteria was met;
    \item \textbf{No}: Not all criteria was met;
    \item \textbf{N/A}: For example, an administrative question for the course.
\end{itemize}

The administrative question category was introduced to help immediately identify data that is not related to programming, and therefore would not be applicable to inclusion in a model fine-tuning set.

Within this dataset, only 528 pairs met all the criteria for inclusion, constituting 21\% of the dataset. Another 53\% contained inappropriate or incorrect responses, deemed not fit for inclusion, and 25\% were not applicable. While this activity was largely helpful in identifying pairs focused on contextualised code queries, it also illuminated some suitability flaws in the initial dataset.

\subsubsection{LLM-Based Data Enhancements}
Once filtered, the dataset was further enhanced by leveraging the OpenAI API with GPT-4o. The existing dataset of 528 pairs suffered from some grammatical issues, as the forum platform does not require grammatical correctness in responses.

The manually filtered dataset was loaded into a Pandas dataframe using a Python script. The content of each cell was used as input to an API call to GPT-4o, and the output replaced the previous cell content. The following system prompt was given to the model, to encourage correct grammatical structure, as well as correct code formatting in the style of general GPT output:

\textit{You are a grammar corrector. Correct the spelling, punctuation and spacing in each cell. Format code snippets with correct spacing and surround by backticks.}

This approach was highly effective in improving the grammatical quality and formal style of the dataset. This is in line with the overarching goal of providing a contextually rich dataset with digestible responses to encourage better student understanding and learning outcomes.

Following all pre-processing steps, the dataset was successfully used to conduct supervised fine-tuning on both the ChatGPT 4o and 4o-mini models to produce two new state-of-art models, GuideLM and GuideLM-mini.

\section{Manual Evaluation Methodology}
Three academics experienced with the CS1 course were asked to evaluate the quality of GuideLM and GuideLM-mini responses, alongside their respective base models. This evaluation methodology is adapted from an evaluation method presented in Taylor et al.~\cite{Taylor2024DccModels}. This method required experts to evaluate the quality of responses of a single model for both compile-time and run-time, based on a number of binary design properties (e.g. Conceptual Accuracy). We applied this method on: GuideLM, GPT-4o, GuideLM-mini and GPT-4o mini.

Four hundred random samples of C error explanations from the tool were randomly selected (200 run-time and 200 compile-time). Each error was constructed into a prompt, comprised of a system prompt encouraging a tutor-like perspective, the full C program code, and any run-time values, if applicable. An exemplar prompt structure is as follows:

\begin{verbatim}
system:content: You are a tutor helping a student.
Do not fix the program. Do not give code.
user:content: This is my C program: <User Code/>
Help me understand this error:
<Compiler error and tool explanation/><Variables/><Call Stack/>
This was the command line: <Command line arguments/>
It was given this input: <Standard input/>
Remember, you are tutor helping a student. Don't write code.
\end{verbatim}

The system prompt was used to generate responses from each of the four models. The three evaluators were asked to rank each of the four responses generated for each respective response between 1 (best) and 4 (worst), as well as evaluate each response as True or False on the design properties listed in Table \ref{tab:my_label}. It was not disclosed to the reviewers which response belonged to which model.

Prior to performing independent evaluations, twelve responses absent from the evaluation set were evaluated together to ensure consensus.

The evaluations were recorded within a spreadsheet with each academic allocated a subset of responses to evaluate. Once evaluated and ranked, a Python script was utilised for aggregation and analysis of the results.

\section{Results}
\begin{figure}[htbp]
    \centering
    \definecolor{CUDBlue}{RGB}{2,114,178}
\definecolor{CUDOrange}{RGB}{230,159,0}
\definecolor{CUDGreen}{RGB}{0,158,115}
\definecolor{CUDPurple}{RGB}{204,121,167}

\begin{tikzpicture}
    \begin{axis}[
        ybar,
        bar width=7pt,
        width=\textwidth,
        height=8cm,
        symbolic x coords={C1, C2, C3, C4, C5, C6, C7, C8, C9},
        xtick=data,
        x tick label style={anchor=center},
        ymin=0, ymax=100,
        ylabel={Average Acceptance},
        enlarge x limits=0.05,
        ymajorgrids=true,
        legend pos=north east,
        legend style={draw=none},
        bar shift auto=0,
        cycle list={
            {CUDBlue, fill=CUDBlue!80},
            {CUDOrange, fill=CUDOrange!80},
            {CUDGreen, fill=CUDGreen!80},
            {CUDPurple, fill=CUDPurple!80}
        }
    ]
    
    \addplot+[bar shift=-10.5pt] table[x=Category,y=4o,col sep=comma] {data.csv};
    \addplot+[bar shift=-3.5pt] table[x=Category,y=4o FT,col sep=comma] {data.csv};
    \addplot+[bar shift=3.5pt] table[x=Category,y=4o mini,col sep=comma] {data.csv};
    \addplot+[bar shift=10.5pt] table[x=Category,y=4o mini FT,col sep=comma] {data.csv};
    
    \legend{4o, 4o FT, 4o mini, 4o mini FT}
    \end{axis}
\end{tikzpicture}
    \caption{Average acceptance rates across models at run-time \textit{(C1: Conceptually Accurate; C2: Inaccuracy Present; C3: Suggestions Correct; C4: Relevant to the Error; C5: Relevant to the Novice; C6: Complete Explanation; C7: Overhelpful; C8: Economy of Words; C9: Socratic Guidance)}.}
    \label{fig:avg_rt_acceptance_rates}
\end{figure}
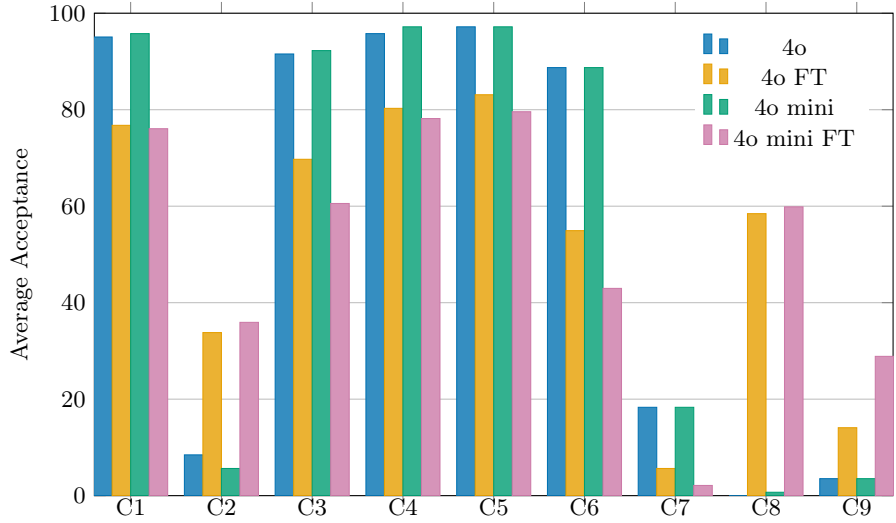

GuideLM and GuideLM-mini significantly outperformed their base model counterparts in the Socratic guidance and economy of words categories, in a consistent trend for both compile-time (CT) and run-time (RT) prompts. Numerical results are presented in \autoref{tab:results}, \autoref{fig:avg_ct_acceptance_rates} and \autoref{fig:avg_rt_acceptance_rates}.


\begin{table}[h]
    \caption{Comparison of fine-tune and base models for compile- and run-time errors.}
    \centering
    \begin{tabular}{|c|c|c|c|c|}\hline
        Category &  CT 4o comp & RT 4o comp & CT mini comp & RT mini comp\\\hline
        Conceptual Accuracy & -9.0\% & -18.3\% & -13.8\% & -19.7\%\\
        Inaccuracy Present & 19.3\% & 25.4\% & 28.3\% & 30.3\%\\
        Suggestions Correct & -11.0\% & -21.8\% & -20.0\% & -31.7\%\\
        Relevant to the Error & -8.3\% & -15.5\% & -10.3\% & -19.0\%\\
        Relevant to the Novice & -11.0\% & -14.1\% & -13.1\% & -17.6\%\\
        Complete Explanation & -17.9\% & -33.8\% & -28.3\% & -45.8\%\\
        Overhelpful & \textbf{-4.8}\% & \textbf{-12.7}\% & \textbf{-31.7}\% & \textbf{-16.2}\%\\
        Economy of Words & \textbf{57.2\%} & \textbf{58.4\%} & \textbf{56.6\%} & \textbf{59.2\%}\\
        Socratic Guidance & \textbf{5.5\%} & \textbf{10.6\%} & \textbf{10.3\%} & \textbf{25.4\%}\\
        \hline
    \end{tabular}

    \label{tab:results}
\end{table}

Across all other categories, the base models outperformed their respective fine-tunes. Both GuideLM models saw between an 8-20\% decrease in conceptual accuracy, error relevance and novice relevance, a 20-45\% decrease in completeness, and 20-30\% increases in inaccuracies present, as presented in \autoref{fig:avg_ct_acceptance_rates} and \autoref{fig:avg_rt_acceptance_rates}. Notably, overhelpfulness was reduced across all fine-tunes at both compile- and run-time by up to 31.7\%. These results align with known qualities of fine-tuning, whereby introduced data reduces overall accuracy \cite{Gekhman2024DoesHallucinations}.

GuideLM models significantly outperformed the base models in the Socratic guidance and economy of words categories, which saw 8\% and 58\% average increases respectively.  This indicates the success of this fine-tuning method to our pedagogical alignment goals, with a definite improvement in making model responses easier to comprehend, and in providing students with questions to consider rather than an explicit solution.

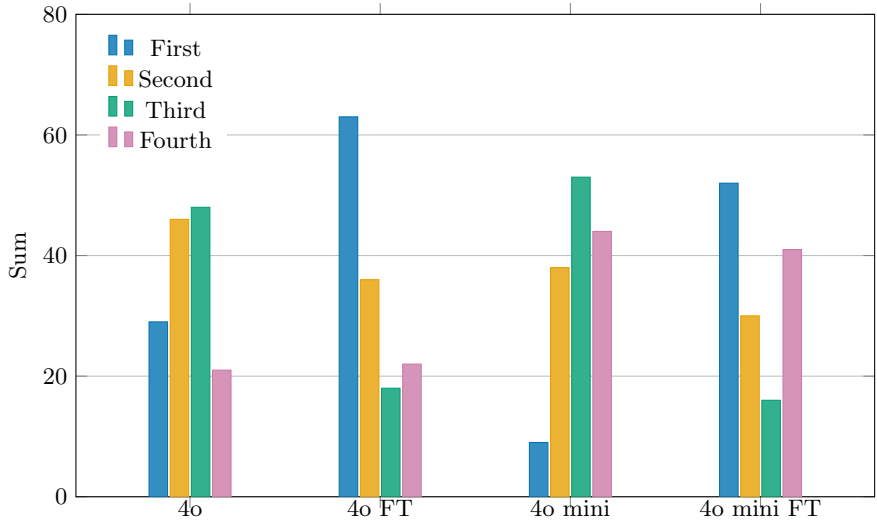
\begin{figure}[htbp]
    \centering
    \definecolor{CUDBlue}{RGB}{2,114,178}
\definecolor{CUDOrange}{RGB}{230,159,0}
\definecolor{CUDGreen}{RGB}{0,158,115}
\definecolor{CUDPurple}{RGB}{204,121,167}

\begin{tikzpicture}
    \begin{axis}[
        ybar,
        bar width=7pt,
        width=\textwidth,
        height=8cm,
        symbolic x coords={4o, 4o FT, 4o mini, 4o mini FT},
        xtick=data,
        x tick label style={anchor=center},
        ymin=0, ymax=80,
        ylabel={Sum},
        enlarge x limits=0.2,
        ymajorgrids=true,
        legend pos=north west,
        legend style={draw=none},
        bar shift auto=0,
        cycle list={
            {CUDBlue, fill=CUDBlue!80},
            {CUDOrange, fill=CUDOrange!80},
            {CUDGreen, fill=CUDGreen!80},
            {CUDPurple, fill=CUDPurple!80},
            {black, fill=black!80}
        }
    ]
    
    \addplot+[bar shift=-12pt] table[x=Model,y=First,col sep=comma] {data_ct_rank.csv};
    \addplot+[bar shift=-4pt] table[x=Model,y=Second,col sep=comma] {data_ct_rank.csv};
    \addplot+[bar shift=4pt] table[x=Model,y=Third,col sep=comma] {data_ct_rank.csv};
    \addplot+[bar shift=12pt] table[x=Model,y=Fourth,col sep=comma] {data_ct_rank.csv};
    
    \legend{First, Second, Third, Fourth}
    \end{axis}
\end{tikzpicture}
    \caption{Average model ranking rates at compile-time.}
    \label{fig:avg_ct_rank_bar}
\end{figure}

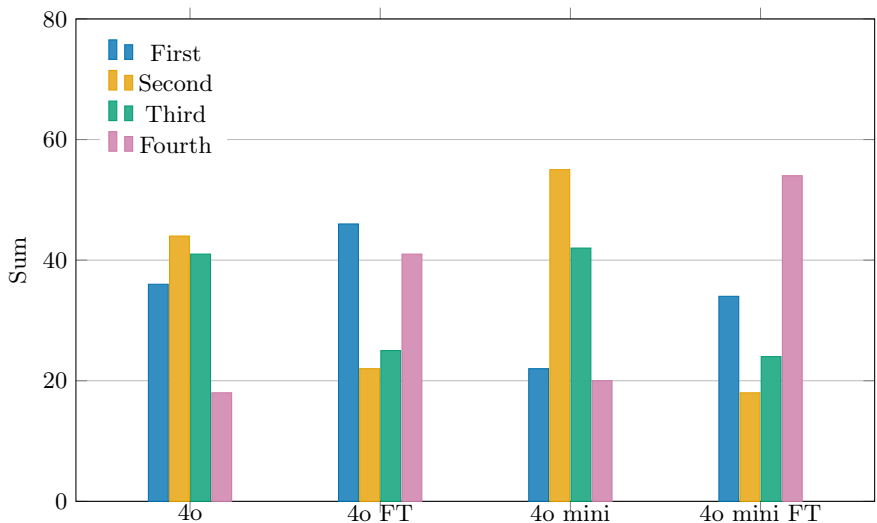
\begin{figure}[htbp]
    \centering
    \definecolor{CUDBlue}{RGB}{2,114,178}
\definecolor{CUDOrange}{RGB}{230,159,0}
\definecolor{CUDGreen}{RGB}{0,158,115}
\definecolor{CUDPurple}{RGB}{204,121,167}

\begin{tikzpicture}
    \begin{axis}[
        ybar,
        bar width=7.5pt,
        width=\textwidth,
        height=8cm,
        symbolic x coords={4o, 4o FT, 4o mini, 4o mini FT},
        xtick=data,
        x tick label style={anchor=center},
        ymin=0, ymax=80,
        ylabel={Sum},
        enlarge x limits=0.2,
        ymajorgrids=true,
        legend pos=north west,
        legend style={draw=none},
        bar shift auto=0,
        cycle list={
            {CUDBlue, fill=CUDBlue!80},
            {CUDOrange, fill=CUDOrange!80},
            {CUDGreen, fill=CUDGreen!80},
            {CUDPurple, fill=CUDPurple!80},
            {black, fill=black!80}
        }
    ]
    
    \addplot+[bar shift=-12pt] table[x=Model,y=First,col sep=comma] {data_rt_rank.csv};
    \addplot+[bar shift=-4pt] table[x=Model,y=Second,col sep=comma] {data_rt_rank.csv};
    \addplot+[bar shift=4pt] table[x=Model,y=Third,col sep=comma] {data_rt_rank.csv};
    \addplot+[bar shift=12pt] table[x=Model,y=Fourth,col sep=comma] {data_rt_rank.csv};
    
    \legend{First, Second, Third, Fourth}
    \end{axis}
\end{tikzpicture}
    \caption{Average model ranking rates at run-time.}
    \label{fig:avg_rt_rank_bar}
\end{figure}

GuideLM model responses were ranked first significantly more often than their base models, indicating successful alignment. For run-time, the fine-tunes were on average 20\% more likely to be ranked first than their base model counterparts, and at least 30\% more likely for compile-time, as presented in \autoref{fig:avg_ct_rank_bar} and \autoref{fig:avg_rt_rank_bar}. GuideLM, based on GPT-4o, was overall most likely to be chosen first from the four models: 32.6\% for run-time and 44.6\% for compile-time. 


\begin{figure}[htbp]
    \centering
    \definecolor{CUDBlue}{RGB}{2,114,178}
\definecolor{CUDOrange}{RGB}{230,159,0}
\definecolor{CUDGreen}{RGB}{0,158,115}
\definecolor{CUDPurple}{RGB}{204,121,167}

\begin{tikzpicture}
    \begin{axis}[
        ybar,
        bar width=7pt,
        width=\textwidth,
        height=8cm,
        symbolic x coords={C1, C2, C3, C4, C5, C6, C7, C8, C9},
        xtick=data,
        x tick label style={anchor=center},
        ymin=0, ymax=100,
        ylabel={Average Acceptance},
        enlarge x limits=0.05,
        ymajorgrids=true,
        legend pos=north east,
        legend style={draw=none},
        bar shift auto=0,
        cycle list={
            {CUDBlue, fill=CUDBlue!80},
            {CUDOrange, fill=CUDOrange!80},
            {CUDGreen, fill=CUDGreen!80},
            {CUDPurple, fill=CUDPurple!80}
        }
    ]
    
    \addplot+[bar shift=-10.5pt] table[x=Category,y=4o,col sep=comma] {data_ct_comp.csv};
    \addplot+[bar shift=-3.5pt] table[x=Category,y=4o FT,col sep=comma] {data_ct_comp.csv};
    \addplot+[bar shift=3.5pt] table[x=Category,y=4o mini,col sep=comma] {data_ct_comp.csv};
    \addplot+[bar shift=10.5pt] table[x=Category,y=4o mini FT,col sep=comma] {data_ct_comp.csv};

    \legend{4o, 4o FT, 4o mini, 4o mini FT}
    \end{axis}
\end{tikzpicture}
    \caption{Average acceptance rates across models at compile-time \textit{(C1: Conceptually Accurate; C2: Inaccuracy Present; C3: Suggestions Correct; C4: Relevant to the Error; C5: Relevant to the Novice; C6: Complete Explanation; C7: Overhelpful; C8: Economy of Words; C9: Socratic Guidance)}.}
    \label{fig:avg_ct_acceptance_rates}
\end{figure}
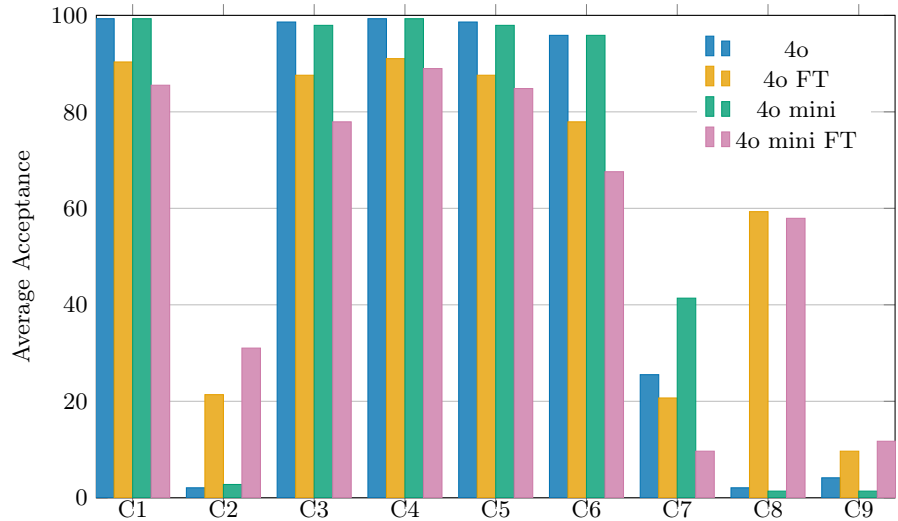

\section{Discussion}
Results indicate that Supervised Fine-Tuning (SFT) is a promising approach towards incorporating pedagogy in large language models (RQ1). While pedagogical goals, such as reducing explicit solutions, producing simpler responses with a Socratic style were also achieved, they came with an overall cost to model accuracy (RQ2). It is yet unclear if this performance cost can be mitigated with improved fine-tuning techniques, reasoning techniques involving Reinforcement Learning, improved foundational models, and datasets, or if the performance trade-off is inherent. We present that while pedagogical alignment comes at the expense of accuracy, the GuideLM models are still preferred by raters and present educational value.

Another aspect to consider is the effort and cost of maintaining fine-tunes as foundational models improve. Since these models are frequently updated, the SFT process must be repeated for each new model.


Ideally, we aim to achieve strong pedagogical performance without directly modifying the foundational models' weights, instead relying on techniques such as RAG and prompting. Therefore, automated benchmarking and evaluation processes, such as those presented in this paper and used by Taylor et al. ~\cite{Taylor2024DccModels}, are crucial for assessing the benefits of the SFT approach and justifying the financial costs of fine-tuning versus other alignment techniques. 

\subsection{Future work}
Results from this study present several avenues for future research and development and the potential to extend to courses beyond CS1. We present how question-answer pair datasets such as those available on course forums can be collected, cleansed, and utilised for supervised fine-tuning of foundational models to develop pedagogically-aligned models, notably in other domains such as introductory mathematics and physics.

GuideLM and GuideLM-mini have been deployed in an A/B testing environment to evaluate its effectiveness in real-world educational settings. This ongoing deployment will provide valuable insights into student learning outcomes, engagement patterns, and the long-term impact of AI-assisted compiler feedback. We are particularly interested in measuring how the increased Socratic guidance affects student problem-solving capabilities and knowledge retention. Ongoing research in this area is critical to evaluate the impact on student learning.

We will also investigate whether broad course forum data from diverse academic disciplines can form effective training datasets for producing general-purpose pedagogical models. Finally, developing pedagogical benchmarks which can automate model evaluation is critical to ensure rapid development cycles.

\section{Conclusion}
This study demonstrates both the potential and limitations of fine-tuning large language models for pedagogical applications in computing education. Our results of an expert analysis comparing language models demonstrate significant improvements in Socratic guidance and economy of words through supervised fine-tuning, with improvements of up to 25.4\% in run-time Socratic guidance of our fine-tuned models (GuideLM, GuideLM-mini). The success of our approach in improving Socratic guidance suggests that fine-tuning can help align AI systems with established pedagogical principles, potentially offering a middle ground between completely automated solutions and traditional human tutoring. However, these gains were accompanied by decreases in conceptual accuracy, highlighting the inherent trade-offs in model specialisation. Despite these trade-offs, expert raters still prefer the GuideLM fine-tunes for pedagogical purposes. 

The integration of GuideLM into the DCC compiler presents a novel approach to providing scalable, personalised support in introductory programming courses. Our approach necessitates deeper engagement in the problem-solving process. This aligns with constructivist learning theories and cognitive load theories that emphasise the importance of guided discovery in education.

Our findings contribute to the broader discourse on AI in education by demonstrating that LLMs can be effectively adapted for specific educational contexts, though this may involve some compromises. Despite the trade-off between pedagogical goals and overall model accuracy, our results suggest that carefully implemented AI systems can serve as valuable tools in computing education, supporting rather than supplanting traditional learning processes.

\printbibliography

\end{document}